\documentclass[letterpaper]{article} 
\usepackage{aaai2027}  
\nocopyright
\usepackage[hyphens]{url}  
\usepackage{graphicx} 
\usepackage{natbib}  
\usepackage{caption} 
\usepackage{algorithm}
\usepackage{algpseudocode}

\usepackage{newfloat}
\usepackage{listings}

\usepackage{amssymb}
\usepackage{amsmath}
\usepackage{colortbl}
\usepackage{xcolor}         

\DeclareCaptionStyle{ruled}{labelfont=normalfont,labelsep=colon,strut=off} 
\floatstyle{ruled}
\newfloat{listing}{tb}{lst}{}
\floatname{listing}{Listing}

\usepackage{booktabs}
\usepackage{multirow,makecell}

\title{ST$^{2}$U: Stateful Test-Time Unlearning via Restricted Knowledge Boundary Control}
\author {
    Xunlei Chen\textsuperscript{\rm 1}
    Qinghui Gong\textsuperscript{\rm 2},
    Ruini Xue\textsuperscript{\rm 1},
    Yaodong Hu\textsuperscript{\rm 1},
    Tian Lan\textsuperscript{\rm 1},
    Wenhong Tian\textsuperscript{\rm 1}\
}

\affiliations {
    \textsuperscript{\rm 1}University of Electronic Science and Technology of China \quad
    \textsuperscript{\rm 2}Southwest Jiaotong University \\
    xunlei@std.uestc.edu.cn
}

\begin{document}

\maketitle

\begin{abstract}
Controlling restricted knowledge in large language models is essential for managing knowledge access in Web-facing services. Test-time unlearning addresses this need without updating model parameters, but successful local correction may not persist as autoregressive generation reconstructs hidden states from the prompt, cache, and generated prefix. We study this gap between local correction and persistent suppression, manifested as restricted knowledge re-entry. To address it, we propose Stateful Test-Time Unlearning (ST$^2$U), which carries correction history across tokens for boundary-aware control. ST$^2$U models restricted knowledge risk in low-dimensional invertible coordinates while preserving the orthogonal residual. During inference, ST$^2$U monitors the evolving trajectory and intervenes when states exceed the learned risk threshold. Boundary-normal descent addresses immediate risk, while tangent-projected historical guidance reuses previous corrections to promote persistent suppression. Contextual anchoring limits representation drift, and a fixed-dimensional state propagates correction history to subsequent positions. Across three benchmarks and three model families, ST$^2$U delivers the strongest overall balance, combining best or second-best retention with competitive forgetting and substantially less restricted knowledge re-entry than test-time baselines (13.76--19.84\% versus 30.44--59.10\%).
\end{abstract}

\section{Introduction}
Large language models (LLMs) increasingly support conversational search and interactive information access on the Web, generating responses that shape how users discover and consume information~\cite{yuan2025query}. However, their large-scale pretraining and fine-tuning corpora may contain hazardous knowledge~\cite{li2024wmdp}, private information~\cite{NEURIPS2025_c9826b9e}, or copyrighted content~\cite{shi2025muse} that can resurface in generated responses. Controlling the reproduction of such knowledge while preserving useful information access is therefore important for trustworthy Web services~\cite{shao2026baldro}. Machine unlearning addresses this need by removing knowledge associated with designated data or concepts while preserving utility on non-target inputs and downstream tasks~\cite{ranjan2026razor}. Ideally, an unlearned model should approximate one retrained without the target data while retaining general language understanding and reasoning capabilities~\cite{Gong_2026_CVPR}. Yet retraining LLMs is prohibitively expensive, and repeated parameter updates can also degrade overall model utility~\cite{yu2025unierase}. Practical methods therefore seek effective forgetting at substantially lower cost while preserving the quality of user-facing responses~\cite{zhang2025price}.

Training-time approximate unlearning remains computationally expensive~\cite{guo2025forgetting,xu2025relearn} and can impair retained utility~\cite{wang2026fupareto}.
Test-time unlearning instead freezes model parameters and intervenes during inference through prompt control~\cite{liu2024large,wang2026cap}, token-distribution adjustment~\cite{liu2025rethinking}, or activation editing~\cite{shen2026llm}.
Decoding- and activation-based methods usually apply isolated corrections at individual decoding positions, without modeling whether those effects persist throughout generation as the autoregressive context evolves.

\begin{figure}[!t]
\centering
\includegraphics[width=1.0\linewidth]{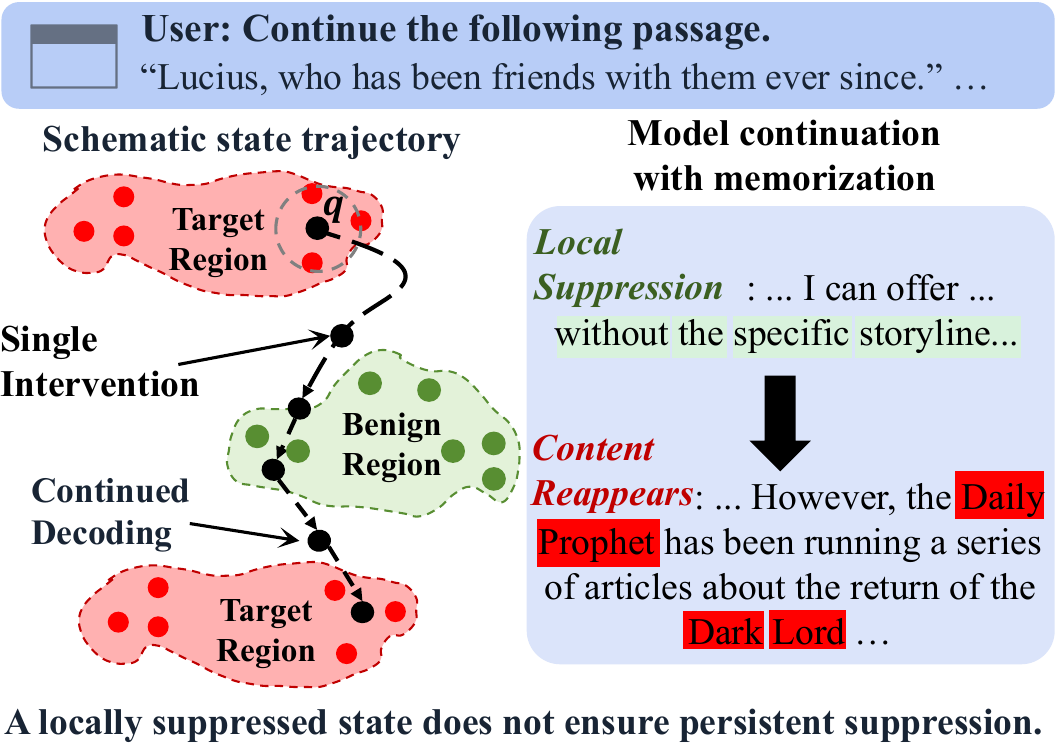}
\caption{Example of restricted knowledge re-entry during decoding. A locally effective intervention initially yields a safe high-level explanation, but later decoding returns to the restricted topic and produces target details.} 
\label{fig:intro}

\end{figure}

We identify an overlooked failure mode in test-time unlearning caused by the lack of temporal persistence in pointwise test-time control.
At each decoding step, autoregressive generation constructs a new hidden state from the prompt, cached context, and generated prefix.
A local edit may correct the current state but imposes no constraint on the subsequent trajectory.
As shown in Fig.~\ref{fig:intro}, later states can therefore return to regions that support restricted generation even after a locally successful intervention.
We call this within-response failure \textit{restricted knowledge re-entry}.

Existing test-time unlearning objectives mainly assess suppression at individual inputs, outputs, or intervention steps~\cite{tutek2025measuring,zhao2025qwen3guard,wang2025towards}. They do not require corrections to remain effective across decoding as the generation context evolves. We therefore formulate stateful test-time unlearning as online control of the hidden-state trajectory. The controller minimizes hidden-state perturbation at each step while keeping risk below the restricted knowledge boundary. The risk constraint enforces forgetting, while minimum perturbation preserves retained knowledge. Persistent test-time unlearning thus requires trajectory-wide risk control across tokens rather than independent pointwise corrections.

To address this problem, inspired by fixed-dimensional recurrent state modules~\cite{pan2025spikingbrain}, we propose \textbf{\underline{S}}tateful \textbf{\underline{T}}est-\textbf{\underline{T}}ime \textbf{\underline{U}}nlearning via restricted knowledge boundary control (\textbf{ST$^{2}$U}), a stateful trajectory-control framework for frozen LLMs powering interactive Web services. Restricted Knowledge Geometry first maps target-related activations into low-dimensional invertible coordinates, estimates a continuous risk boundary using density ratios, and preserves the orthogonal complement. Contextual sanitized references further specify how generation can continue coherently. Stateful Boundary Control then monitors risk throughout generation, moves risky states below the boundary through normal descent, and anchors corrections to context-compatible references to limit semantic drift. A fixed-dimensional state summarizes previous corrections and acts only within the tangent space of the current boundary, maintaining temporal consistency without opposing risk reduction.
Across three benchmarks, ST$^{2}$U achieves leading overall trade-offs: its lowest forgetting rate on each benchmark lies between 80.43\% and 90.17\%, and it retains 90.92\% of the original MMLU accuracy on average across benchmarks.
It limits restricted knowledge re-entry to 13.76--19.84\%, versus 30.44--59.10\% for test-time baselines.
Our contributions are:

\begin{itemize}
    \item  We construct a restricted knowledge risk geometry that represents contextual trajectories in invertible low-dimensional coordinates while preserving activation components orthogonal to the target knowledge.   
    
    \item We propose ST$^{2}$U, which combines normal--tangential boundary control, contextual anchoring, and a fixed-dimensional correction state to control hidden-state trajectories throughout generation.
    
    \item We validate ST$^{2}$U across WMDP, RWKU, and MUSE-Books, demonstrating the best overall forget--retain trade-offs among the compared methods, substantially lower restricted knowledge re-entry rates, and stable effectiveness across diverse model families.
\end{itemize}

\section{Related Work}

\textbf{Parameter-based Unlearning.}
Gradient Ascent (GA)~\cite{thudi2022unrolling} maximizes forget loss, but drives the model toward degenerate outputs. Representation Misdirection for Unlearning (RMU)~\cite{li2024wmdp} pushes forget-set activations toward a fixed random target vector while regularizing retain-set activations toward those of the frozen model. AS~\cite{tan2026wisdom} weakens lexical and semantic associations through attention smoothing and self-distillation, while ALTER~\cite{chen2026alter} combines token entropy with asymmetric LoRA to separate forgetting from retention. These methods improve the stability and granularity of parameter-based unlearning~\cite{li2026cipo,li2026cross}, but encode forgetting in updated weights without an online mechanism for context-dependent control over successive decoding steps.

\textbf{Test-Time Unlearning.}
Test-time unlearning methods operate through input, agentic, decoding, and activation interfaces. SPUL~\cite{bhaila2025soft} learns soft prompts that induce forgetting at the input, while ALU~\cite{sanyal2025agents} coordinates specialized agents to process unlearning requests. During decoding, SEGUE~\cite{pu2026decoding} detects forget-related queries and suppresses factual units through entropy-guided decoding. 
Activation-based methods directly edit hidden representations through static~\cite{cao2025scans}, input-conditioned~\cite{lee2026direct,pmlr-v306-sun26d}, or nonlinear steering~\cite{nguyen2026beyond}.

Existing test-time methods control a query, token distribution, or current activation, but do not explicitly model restricted knowledge risk over the evolving hidden-state trajectory. Their interventions may be locally selective, yet remain independent across decoding steps and carry no correction history. ST$^2$U targets this missing temporal coupling by treating test-time unlearning as stateful trajectory control rather than a sequence of isolated corrections.

\section{Problem Formulation}
\subsection{Restricted Knowledge Re-entry}
Let the frozen LLM be $f_\theta$. Given a prompt $q=(x_1,\ldots,x_n)$, the model autoregressively generates future tokens $x_{n+1:n+T}$. At position $t$, let $\mathbf h_t\in\mathbb R^d$ be the pre-control hidden state at layer $\ell$ and $\mathbf h_t'$ its controlled counterpart. We consider a task-specific forget set $\mathcal F$ for targeted unlearning.

\begin{figure}[!ht]
\centering
\includegraphics[width=1.0\linewidth]{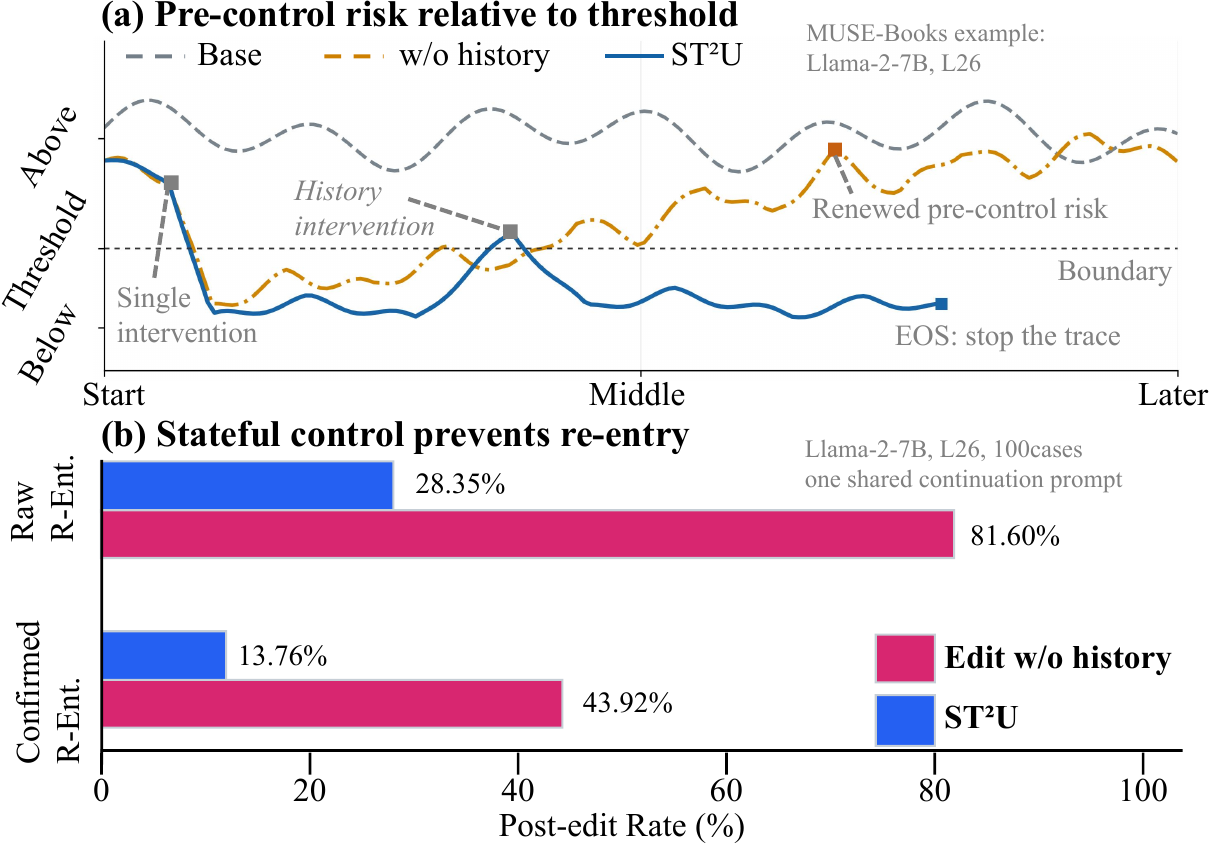}
\caption{Restricted knowledge re-entry during generation on MUSE-Books. (a) Pre-control risk trajectories illustrate how risk can rise again after a local intervention. (b) Geometric and LLM-confirmed re-entry rates compare ST$^2$U with its history-free variant.}
\label{fig:obs}

\end{figure}

A stateless pointwise editor may intervene at every decoding step, but does not carry correction history across tokens. As generation proceeds, subsequent states depend on the evolving context, including the prompt, cached representations, and generated prefix. A successful local correction therefore does not guarantee that later pre-control states remain below the restricted knowledge boundary. Figure~\ref{fig:obs} illustrates this issue through a long-form continuation, relating the risk trajectory to target-text generation. We define this recurrence as restricted knowledge re-entry:
\begin{equation}
R_{\mathcal F}(\mathbf h_{t_0}')\leq\tau,\qquad
\exists\,t_1>t_0:
R_{\mathcal F}(\mathbf h_{t_1})>\tau.
\label{eq:reentry}
\end{equation}

Here, \(R_{\mathcal F}\) assigns each hidden state a continuous restricted knowledge risk score, and \(\tau\) is its threshold. Eq.~\ref{eq:reentry} compares a corrected state satisfying the risk constraint at \(t_0\) with a later violating state before control at \(t_1\). Re-entry means a violation recurs after successful correction, even if the new violation is later corrected. Stateful test-time unlearning fixes current violations and uses past corrections to reduce recurrence during later decoding.

\subsection{Stateful Test-Time Unlearning}
With $\theta$ frozen, let $\mathbf h_t(\Delta\mathbf h_{<t})$ denote the pre-control state induced by the generation history and earlier corrections. We seek minimum corrections that keep every controlled prefill and decoding state outside the risk region:

\begin{equation}
\begin{split}
\{\Delta\mathbf h_t^*\}
&=
\arg\min_{\{\Delta\mathbf h_t\}}
\sum_{t=1}^{n+T}
\big\|\Delta\mathbf h_t\big\|_2^2, \\
&\text{s.t.}\quad
R_{\mathcal F}\!\left(
\mathbf h_t(\Delta\mathbf h_{<t})+\Delta\mathbf h_t
\right)
\leq \tau,\quad \forall t.
\end{split}
\label{eq:trajectory_objective}
\end{equation}

The post-control state is $\mathbf h_t'=\mathbf h_t(\Delta\mathbf h_{<t})+\Delta\mathbf h_t$. Its dependence on earlier corrections couples the per-position constraints along the generated trajectory. During decoding, this dependence is mediated by the controlled generation history. Batched prefill computes transformer activations in parallel, so ST$^2$U approximates the causal dependence by scanning prompt positions to propagate its controller state without recomputing later prefill activations. Eq.~\ref{eq:trajectory_objective} expresses forgetting through the risk constraint and retention through minimum perturbation. Our controller restricts each correction to a learned subspace of normalized activations and provides a tractable local surrogate for this trajectory objective. We target deployment-time suppression of target-knowledge access and expression, not parameter-level data deletion.

\section{ST$^2$U: Stateful Boundary Control \\ for Test-Time Knowledge Unlearning}

\begin{figure*}[!ht]
\centering
\includegraphics[width=0.95\linewidth]{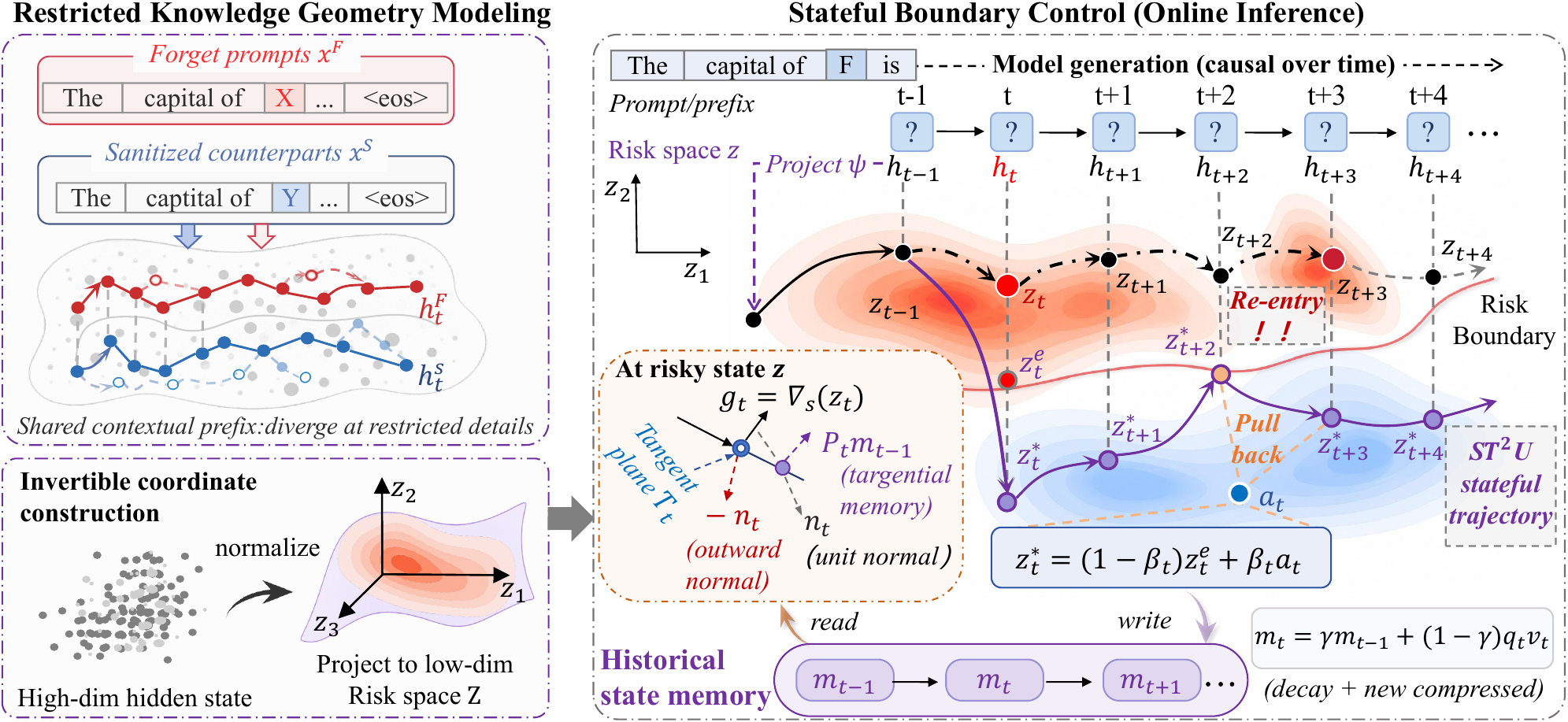}
\caption{Overview of ST$^{2}$U. Offline paired trajectories define low-dimensional invertible coordinates, KDE-based risk geometry, and context-dependent references. During prefill and decoding, risk-triggered control combines normal descent with tangent-projected history, steps toward the risk boundary, and anchors feasible states to local references. A fixed-dimensional state propagates accepted correction directions across tokens to mitigate restricted knowledge re-entry.} 
\label{fig:method}

\end{figure*}

ST$^2$U comprises two components (Fig.~\ref{fig:method}). Restricted Knowledge Geometry (RKG) constructs a task-specific, low-dimensional risk geometry from paired restricted and sanitized trajectories before deployment. Stateful Boundary Control (SBC) monitors this geometry during inference and summarizes accepted correction directions in a fixed-dimensional state $\mathbf m_t\in\mathbb R^k$, where $k\ll d$.

\subsection{Restricted Knowledge Geometry Modeling}
\paragraph{Paired contextual trajectories.}
For each restricted sequence $x_i^F\in\mathcal F$, we construct an offline paired reference $x_i^S$ that preserves shared high-level context and non-target content but omits target-specific details. The pair supplies contrastive supervision for isolating target-related activation variations from shared contextual structure. The pairs are used to learn a transferable representation geometry, not to prescribe a response tied to one offline sequence, so the frozen model can still generate under unseen prefixes. At layer $\ell$, we collect activations from the paired sequences. The reference construction procedure is described in Appendix~A.

\paragraph{Low-dimensional invertible coordinates.}
Let $\boldsymbol\mu$ and $\boldsymbol\sigma$ be activation statistics estimated offline, and normalize each state as $\widetilde{\mathbf h}=(\mathbf h-\boldsymbol\mu)\oslash\boldsymbol\sigma$. From aligned trajectory differences, we learn a column-orthonormal basis $\mathbf U\in\mathbb R^{d\times k}$. We project the normalized state into target-related coordinates $\mathbf p$ and their orthogonal residual $\mathbf r$, then transform $\mathbf p$ with an invertible map $\psi:\mathbb R^k\rightarrow\mathbb R^k$:

\begin{equation}
\begin{aligned}
\mathbf p &= \mathbf U^\top\widetilde{\mathbf h},
&
\mathbf r &= (\mathbf I-\mathbf U\mathbf U^\top)\widetilde{\mathbf h},
\\
\mathbf z &= \psi(\mathbf p),
&
\widetilde{\mathbf h} &= \mathbf r+\mathbf U\psi^{-1}(\mathbf z).
\end{aligned}
\end{equation}

The basis $\mathbf U$ is designed to capture low-dimensional variation associated with restricted knowledge, while $\mathbf r$ retains components orthogonal to that variation. The map $\psi$ changes coordinates but does not itself perform unlearning. Because the same invertible map is applied to both distributions, its Jacobian cancels from their ideal density ratio. In finite samples, $\psi$ regularizes local scale and curvature to improve KDE conditioning and control-gradient stability.
Together, the orthonormal projection and invertible map define an exact decomposition $\widetilde{\mathbf h}\leftrightarrow(\mathbf r,\mathbf z)$. ST$^2$U edits only $\mathbf z$ and restores the unchanged residual $\mathbf r$ during reconstruction. The invertible architecture is specified in Appendix, and coordinate preservation is established in Appendix.

\paragraph{Risk geometry and contextual references.}
In $\mathbf z$-space, kernel density estimation (KDE)~\cite{scott2015multivariate} models restricted and sanitized trajectories with densities $d_F$ and $d_S$. Their log-density ratio defines restricted knowledge risk:
\begin{equation}
\begin{aligned}
s(\mathbf z)
=\log d_F(\mathbf z)-\log d_S(\mathbf z),
\quad
R_{\mathcal F}(\mathbf h)
=s\!\left(
\psi\!\left(\mathbf U^\top\widetilde{\mathbf h}\right)
\right).
\end{aligned}
\label{eq:risk_geometry}
\end{equation}

A larger $s(\mathbf z)$ means that restricted trajectories provide stronger local support than sanitized trajectories. The level set $s(\mathbf z)=\tau$ defines the task-specific boundary, and states with $s(\mathbf z)>\tau$ form the risk region. Compared with $d_F$ alone, the density ratio reduces sensitivity to regions supported by both distributions.
For local anchoring, we group restricted trajectories into $J$ neighborhoods and associate each neighborhood $j$ with a $\mathbf z$-space reference $\mathbf a_j^S\in\mathbb R^k$ estimated from its paired trajectories. Let $\rho_j^F(\mathbf z)$ denote the normalized KDE responsibility of neighborhood $j$, with $\sum_{j=1}^{J}\rho_j^F(\mathbf z)=1$. The contextual reference is
\begin{equation}
\mathbf a(\mathbf z)
=\sum_{j=1}^{J}
\rho_j^F(\mathbf z)\mathbf a_j^S.
\label{eq:sanitized_reference}
\end{equation}

This responsibility-weighted statistic adapts supervision to the current activation. The coordinate-training objective is given in the Appendix, and multi-target risk routing is discussed in the Appendix.

\subsection{Stateful Boundary Control}
\paragraph{Risk-triggered control.}
At position $t$, the controller maps $\mathbf h_t$ to $\mathbf z_t$ and evaluates $s_t=s(\mathbf z_t)$. The historical state starts at $\mathbf m_0=\mathbf 0$. If $s_t\leq\tau$, the hidden state is unchanged and the history decays as $\mathbf m_t=\gamma\mathbf m_{t-1}$. If $s_t>\tau$, the controller applies the boundary correction described below. The same test is used throughout prefill and decoding. The complete online inference procedure is provided in Algorithm of Appendix.

\paragraph{Normal descent with tangential state.}
Let $\mathbf g_t=\nabla s(\mathbf z_t)$. For $s_t>\tau$ and $\|\mathbf g_t\|_2>\epsilon_g$, its normalized direction is the local boundary normal, and the orthogonal projector defines the tangent space:
\begin{equation}
\begin{aligned}
\mathbf n_t
&=\frac{\mathbf g_t}{\|\mathbf g_t\|_2},
\quad
\mathbf P_t
=\mathbf I-\mathbf n_t\mathbf n_t^\top,
\\
\mathbf d_t
&=\operatorname{Normalize}\!\left(
-\mathbf n_t+
\lambda_m\mathbf P_t\mathbf m_{t-1}
\right).
\end{aligned}
\label{eq:control_direction}
\end{equation}

The normal term $-\mathbf n_t$ decreases current risk, whereas $\mathbf P_t\mathbf m_{t-1}$ retains the component of historical guidance tangent to the current level set. Because $\mathbf g_t^\top\mathbf P_t\mathbf m_{t-1}=0$, history cannot reverse the normal descent, although normalization reduces its normal magnitude. Thus, $\mathbf g_t^\top\mathbf d_t<0$ whenever the gradient is nonzero. The low-gradient fallback and its risk-monotonicity property are established in Appendix.
The normal--tangential direction corresponds to the following local surrogate:
\begin{equation}
\begin{aligned}
\Delta\mathbf z_t^*
&=\arg\min_{\Delta\mathbf z}\;
\frac{1}{2}\|\Delta\mathbf z\|_2^2
-\lambda_t
\left\langle
\mathbf P_t\mathbf m_{t-1},
\Delta\mathbf z
\right\rangle,
\\
\text{s.t.}\quad&
s_t+\mathbf g_t^\top\Delta\mathbf z\leq\tau,
\qquad
\lambda_t=
\lambda_m\frac{[s_t-\tau]_+}{\|\mathbf g_t\|_2}.
\end{aligned}
\label{eq:local_surrogate}
\end{equation}

The quadratic term penalizes intervention magnitude in $\mathbf z$-space, the linear term favors consistent historical tangential alignment, and the linearized constraint actively reduces risk. We compute the step along $\mathbf d_t$ as
\begin{equation}
\begin{aligned}
\alpha_t
&=\operatorname{clip}\!\left(
\frac{[s_t-\tau]_+}
{-\mathbf g_t^\top\mathbf d_t+\epsilon},
0,\alpha_{\max}
\right).
\end{aligned}
\label{eq:boundary_step}
\end{equation}

Ignoring clipping and $\epsilon$, the update $\alpha_t\mathbf d_t$ exactly solves Eq.~\ref{eq:local_surrogate}. Because $s$ is nonlinear, we set $\mathbf z\leftarrow\mathbf z+\alpha_t\mathbf d_t$, recompute the risk and gradient, and repeat until the boundary is reached or $K_{\max}$ iterations are used. This provides a tractable local, per-token surrogate for Eq.~\ref{eq:trajectory_objective} without expensive online parameter optimization.

\begin{table*}[h]
\centering
\caption{Main results on RWKU and WMDP. \textsuperscript{$\dagger$}, \textsuperscript{$\ddagger$}, and \textsuperscript{$\star$} denote parameter-updating, in-context, and test-time intervention methods. R-Ent. denotes confirmed re-entry. Base\textsuperscript{\rm E} has no intervention; we report the occurrence of target domain knowledge. $\pm$ denotes the standard deviation over three seeds (42/12345/100) for operations during offline construction or training.}
\resizebox{1.0\textwidth}{!}{
\begin{tabular}{@{}lccccc|cccc@{}}
\toprule
\multirow{2}{*}{\textbf{Method}}
& \multicolumn{5}{c}{\textbf{RWKU}}
& \multicolumn{4}{c}{\textbf{WMDP}} \\
\cmidrule(lr){2-6}\cmidrule(lr){7-10}
& \textbf{FB (\%, $\downarrow$)}
& \textbf{QA (\%, $\downarrow$)}
& \textbf{Gen. (\%, $\uparrow$)}
& \textbf{Flu. (1--5, $\uparrow$)}
& \textbf{R-Ent. (\%, $\downarrow$)}
& \textbf{Bio. (\%, $\downarrow$)}
& \textbf{Cyber (\%, $\downarrow$)}
& \textbf{Gen. (\%, $\uparrow$)}
& \textbf{R-Ent. (\%, $\downarrow$)} \\
\midrule
\rowcolor{gray!20} \multicolumn{10}{c}{\textit{\textbf{Llama-3.1-8B-Instruct}}} \\
\midrule
Base
& 88.31 & 74.92 & 64.38 & 4.20 & 89.14\textsuperscript{\rm E} & 72.74 & 47.35 & 64.38 & 91.14\textsuperscript{\rm E} \\
RMU\textsuperscript{$\dagger$}
& 33.10$\pm$0.20 & 27.84$\pm$0.10 & 56.71$\pm$0.30 & 2.90$\pm$0.21 & \underline{21.30$\pm$0.29} & 35.68$\pm$0.22 & 38.82$\pm$0.35 & 51.31$\pm$0.14 & \underline{19.80$\pm$0.12} \\
AS\textsuperscript{$\dagger$}
& \textbf{10.90$\pm$0.16} & \textbf{14.62$\pm$0.15} & 57.40$\pm$0.26 & 3.43$\pm$0.25 & 51.80$\pm$0.16 & \textbf{27.72$\pm$0.30} & 33.82$\pm$0.23 & 57.74$\pm$0.23 & 49.40$\pm$0.20 \\
ICUL\textsuperscript{$\ddagger$}
& 50.20$\pm$0.00 & 43.77$\pm$0.00 & \underline{60.80$\pm$0.00} & \textbf{4.07$\pm$0.00} & 83.50$\pm$0.00 & 55.90$\pm$0.00 & 44.30$\pm$0.00 & \underline{59.60$\pm$0.00} & 82.80$\pm$0.00 \\
ULD\textsuperscript{$\star$}
& 32.40$\pm$0.17 & 30.94$\pm$0.24 & 56.70$\pm$0.27 & 3.36$\pm$0.27 & 50.20$\pm$0.30 & 35.90$\pm$0.24 & 35.10$\pm$0.26 & 55.50$\pm$0.17 & 48.40$\pm$0.36 \\
S{\small CANS}\textsuperscript{$\star$}
& 22.80$\pm$0.36 & 28.47$\pm$0.26 & 58.40$\pm$0.25 & 3.61$\pm$0.28 & 54.70$\pm$0.36 & 31.70$\pm$0.17 & 34.00$\pm$0.26 & 57.20$\pm$0.24 & 53.10$\pm$0.25 \\
I{\small NNSTEER}\textsuperscript{$\star$}
& 20.60$\pm$0.20 & 24.91$\pm$0.27 & 59.80$\pm$0.22 & 3.79$\pm$0.19 & 48.90$\pm$0.25 & 30.60$\pm$0.23 & 32.80$\pm$0.25 & 58.60$\pm$0.20 & 47.20$\pm$0.24 \\
S{\small EGUE}\textsuperscript{$\star$}
& 17.18$\pm$0.30 & 20.94$\pm$0.40 & 59.00$\pm$0.26 & 3.68$\pm$0.12 & 30.84$\pm$0.64 & 31.78$\pm$0.22 & \underline{32.44$\pm$0.28} & 58.42$\pm$0.24 & 32.28$\pm$0.36 \\
ST$^2$U (Ours)\textsuperscript{$\star$}
& \underline{13.40$\pm$0.33} & \underline{17.08$\pm$0.35} & \textbf{60.90$\pm$0.19} & \underline{3.93$\pm$0.29} & \textbf{19.20$\pm$0.22} & \underline{28.04$\pm$0.20} & \textbf{30.48$\pm$0.39} & \textbf{60.58$\pm$0.14} & \textbf{13.76$\pm$0.31} \\
\midrule
\rowcolor{gray!20} \multicolumn{10}{c}{\textit{\textbf{Qwen3-14B}}} \\
\midrule
Base
& 63.06 & 45.42 & 79.35 & 4.41 & 92.02\textsuperscript{\rm E} & 75.51 & 61.06 & 79.35 & 94.52\textsuperscript{\rm E} \\
ULD\textsuperscript{$\star$}
& 25.80$\pm$0.25 & 26.94$\pm$0.37 & 66.00$\pm$0.34 & 3.48$\pm$0.26 & 50.10$\pm$0.17 & 37.80$\pm$0.22 & 39.20$\pm$0.29 & 64.80$\pm$0.19 & 48.20$\pm$0.34 \\
S{\small CANS}\textsuperscript{$\star$}
& 21.50$\pm$0.27 & 28.11$\pm$0.18 & \underline{68.70$\pm$0.18} & \underline{3.79$\pm$0.26} & 56.40$\pm$0.25 & 34.60$\pm$0.28 & 35.80$\pm$0.19 & \underline{67.50$\pm$0.27} & 54.10$\pm$0.32 \\
S{\small EGUE}\textsuperscript{$\star$}
& \underline{16.50$\pm$0.22} & \underline{24.10$\pm$0.14} & 66.46$\pm$0.24 & 3.68$\pm$0.06 & \underline{30.44$\pm$0.14} & \underline{30.60$\pm$0.20} & \underline{33.94$\pm$0.16} & 66.58$\pm$0.08 & \underline{34.14$\pm$0.40} \\
ST$^2$U (Ours)\textsuperscript{$\star$}
& \textbf{12.10$\pm$0.23} & \textbf{19.57$\pm$0.17} & \textbf{70.62$\pm$0.18} & \textbf{4.10$\pm$0.24} & \textbf{19.84$\pm$0.12} & \textbf{30.14$\pm$0.27} & \textbf{32.37$\pm$0.25} & \textbf{69.42$\pm$0.15} & \textbf{16.20$\pm$0.17} \\
\midrule
\rowcolor{gray!20} \multicolumn{10}{c}{\textit{\textbf{SpikingBrain-7B}}} \\
\midrule
Base
& 80.10 & 67.40 & 65.97 & 4.12 & 90.40\textsuperscript{\rm E} & 68.74 & 44.94 & 65.97 & 91.00\textsuperscript{\rm E} \\
ULD\textsuperscript{$\star$}
& 33.40$\pm$0.18 & 34.85$\pm$0.18 & 56.40$\pm$0.32 & 3.27$\pm$0.21 & 51.00$\pm$0.22 & 45.80$\pm$0.27 & 38.90$\pm$0.26 & 55.20$\pm$0.28 & 46.50$\pm$0.25 \\
S{\small CANS}\textsuperscript{$\star$}
& 28.70$\pm$0.38 & 36.12$\pm$0.29 & \underline{58.60$\pm$0.20} & \underline{3.53$\pm$0.19} & 59.10$\pm$0.27 & 39.80$\pm$0.09 & 36.40$\pm$0.19 & \underline{57.40$\pm$0.19} & 49.80$\pm$0.14 \\
S{\small EGUE}\textsuperscript{$\star$}
& \underline{18.62$\pm$0.38} & \underline{22.80$\pm$0.42} & 58.08$\pm$0.30 & 3.52$\pm$0.16 & \underline{36.28$\pm$0.10} & \underline{34.72$\pm$0.20} & \underline{36.08$\pm$0.30} & 55.50$\pm$0.28 & \underline{37.28$\pm$0.36} \\
ST$^2$U (Ours)\textsuperscript{$\star$}
& \textbf{14.42$\pm$0.20} & \textbf{16.76$\pm$0.38} & \textbf{59.24$\pm$0.25} & \textbf{3.84$\pm$0.20} & \textbf{18.10$\pm$0.27} & \textbf{25.93$\pm$0.15} & \textbf{31.78$\pm$0.24} & \textbf{59.04$\pm$0.25} & \textbf{15.40$\pm$0.18} \\
\bottomrule
\end{tabular}
}
\label{tab:main_rwku_wmdp}
\end{table*}

\paragraph{Contextual anchoring and state update.}
Risk descent may enter low-density regions with limited linguistic support. We therefore evaluate the contextual reference $\mathbf a_t=\mathbf a(\mathbf z_t)$ at the pre-control coordinate and keep it fixed during correction. After obtaining the boundary-corrected state $\mathbf z_t^e$, we softly anchor it to this reference:
\begin{equation}
\begin{aligned}
\mathbf a_t
&=\mathbf a(\mathbf z_t),
\quad
&
\mathbf z_t^*
&=(1-\beta_t)\mathbf z_t^e+\beta_t\mathbf a_t,
\\
\mathbf p_t'
&=\psi^{-1}(\mathbf z_t^*),
&
\mathbf h_t'
&=
\left(\mathbf r_t+\mathbf U\mathbf p_t'\right)
\odot\boldsymbol\sigma+\boldsymbol\mu.
\end{aligned}
\label{eq:anchoring_reconstruction}
\end{equation}

The controller selects $0\leq\beta_t\leq\beta_{\max}$ via finite backtracking. An anchored state is accepted only if it remains below the risk boundary with sufficient support under $d_S$. Otherwise, attraction is reduced or removed. Eq.~\ref{eq:anchoring_reconstruction} updates only the learned target coordinates and restores the residual $\mathbf r_t$ unchanged. Residual preservation and contextual anchoring limit semantic drift, while the historical state below provides temporal consistency.
Finally, the correction direction is compressed into a fixed-dimensional state:

\begin{equation}
\begin{aligned}
\mathbf v_t
&=
\frac{\mathbf z_t^*-\mathbf z_t}
{\|\mathbf z_t^*-\mathbf z_t\|_2+\epsilon},
\quad
q_t
=1-\exp\!\left(
-\frac{[s_t-\tau]_+}{T_g}
\right),
\\
\mathbf m_t
&=
\gamma\mathbf m_{t-1}
+(1-\gamma)q_t\mathbf v_t.
\end{aligned}
\label{eq:state_update}
\end{equation}

The gate $q_t$ increases with the original boundary violation and approaches zero as $s_t\rightarrow\tau^+$. Consecutive high-risk corrections therefore accumulate consistent directional information to discourage subsequent restricted knowledge re-entry across later decoding positions, while low-risk positions decay the history. During prefill, ST$^2$U scans positions causally to initialize $\mathbf m_n$. Decoding continues the same state recurrence without storing full activation histories. The online inference procedure is summarized in Appendix.

\section{Experiments}

\subsection{Experimental Settings}

\noindent$\bullet$ \textbf{Datasets.}
We evaluate ST$^2$U on three benchmarks.
RWKU evaluates the removal of real-world entity knowledge~\cite{cao2024rwku}.
WMDP measures hazardous knowledge in biosecurity (bio) and cybersecurity (cyber)~\cite{li2024wmdp}.
MUSE-Books evaluates copyright unlearning on the Harry Potter corpus~\cite{shi2025muse}.
MMLU measures the preservation of general knowledge~\cite{hendrycks2020measuring}.

\noindent$\bullet$ \textbf{Evaluation Metrics.} For RWKU, we report ROUGE-L recall on fill-in-the-blank (FB) and question-answering (QA) probes. For WMDP, we report Bio/Cyber accuracy, with 25\% as the random-choice reference. For MUSE-Books, we report BLEU and ROUGE-L. General knowledge retention is measured by MMLU accuracy (Gen., \%). GPT-4o fluency~\cite{xu2025obliviate} (Flu., 1--5) measures linguistic fluency and coherence, without assessing factual correctness. For summary statistics, the forgetting rate is 100 minus the forget score, normalized on WMDP by the 25\% chance level as $(100-\mathrm{acc})/75$, and retention is MMLU accuracy relative to the original model.

We additionally report confirmed re-entry (R-Ent.) on RWKU and WMDP. For test-time intervention methods, R-Ent. is the proportion of generation probes in which target knowledge reappears after an observed successful local suppression. For methods without token-level runtime correction (parameter-updating methods and ICUL), the reported value instead measures residual target-knowledge disclosure. For Base, it measures unconstrained disclosure. All rates use the same request-level denominator, maximum generation length, and output-confirmation criterion. Unlike these output-confirmed rates, Eq.~\ref{eq:reentry} defines geometric re-entry through a later pre-control risk violation, which is analyzed separately in Fig.~\ref{fig:obs}(b).

\noindent$\bullet$ \textbf{Baselines.}
We compare ST$^{2}$U with RMU~\cite{li2024wmdp}, ICUL~\cite{pawelczyk2024context},
WHP~\cite{eldan2023s},
Attention-Shifting (AS)~\cite{tan2026wisdom}, ULD~\cite{ji2024reversing}, SEGUE~\cite{pu2026decoding}, and adapted SCANS~\cite{cao2025scans} and INNSTEER~\cite{nguyen2026beyond}.
For RWKU and WMDP, we evaluate Llama-3.1-8B-Instruct~\cite{grattafiori2024llama}, Qwen3-14B~\cite{yang2025qwen3}, and SpikingBrain-7B~\cite{pan2025spikingbrain}. For MUSE-Books, we use Llama-2-7B in place of the ICLM-7B used in the original benchmark~\cite{shi2025muse}.
The ethics statement is provided in Appendix.

\noindent$\bullet$ \textbf{Implementation Details.}
Across all backbones, we use a risk dimension of $k=32$, $J=16$ contextual neighborhoods, and six affine-coupling blocks with hidden width 128. We fit the invertible map for 500 steps using AdamW (learning rate $10^{-3}$, batch size 256). Corrections operate in normalized coordinates with threshold quantile $q_\tau=94\%$ and maximum step $\alpha_{\max}=12$. Geometry is constructed separately for each backbone, while controller hyperparameters remain shared. Restricted and sanitized densities use Gaussian KDEs with Scott's bandwidth rule. We allow at most $K_{\max}=4$ boundary iterations per position. NVIDIA RTX 4090 GPUs x 4. Further implementation details are in Appendix.

\subsection{Main Results}
\noindent$\bullet$ \textbf{Forget--Retain Trade-off.} Table~\ref{tab:main_rwku_wmdp} reports entity forgetting on RWKU (FB and QA), domain forgetting on WMDP (Bio and Cyber accuracy), and retained general capabilities. Table~\ref{tab:main_muse_books} evaluates long-form copyright continuation on MUSE-Books using Llama-2-7B. AS and SEGUE serve as the main reference parameter-updating and test-time baselines, respectively. On Llama-3.1-8B, AS achieves the lowest FB, QA, and Bio scores, whereas ST\(^{2}\)U retains 3.50 and 2.84 more MMLU points on RWKU and WMDP, respectively, with lower Cyber accuracy. Although SEGUE offers a strong trade-off, ST\(^{2}\)U improves forgetting and retention across all three backbones. On Llama-3.1-8B, ST\(^{2}\)U reduces FB and QA by 3.78 and 3.86 points relative to SEGUE, while improving retained MMLU by 1.90 and 2.16 points on RWKU and WMDP, respectively. It also reduces R-Ent. by 11.64 and 18.52 points, indicating more persistent suppression. On MUSE, ST\(^{2}\)U approaches AS in BLEU (14.61 vs.\ 13.72) while retaining 4.78 more MMLU points and higher fluency. ICUL maintains high fluency but achieves weaker target forgetting.

\noindent$\bullet$ \textbf{Model Transferability.} ST\(^{2}\)U consistently outperforms the evaluated test-time intervention baselines on all reported RWKU and WMDP metrics across Llama-3.1-8B, Qwen3-14B, and SpikingBrain-7B. Because the geometry is built in each backbone's own representation space, ST\(^{2}\)U does not require representations to be aligned across models. These results support applying the same control procedure to different backbones, not zero-shot transfer of a fitted controller. The MUSE results on Llama-2-7B further extend the evaluation to long-form continuation. 
More cross-model results are presented in Appendix.

\begin{table}[t]
\centering
\caption{Copyright unlearning results on MUSE-Books.}
\resizebox{\columnwidth}{!}{
\begin{tabular*}{\columnwidth}{
@{\extracolsep{\fill}}lcccc@{}
}
\toprule
\multirow{2}{*}{\textbf{Method}}
& \multicolumn{2}{c}{Forget Perf.}
& \multicolumn{2}{c}{Retain Perf.} \\
\cmidrule(lr){2-3}
\cmidrule(lr){4-5}
& {BLEU$\downarrow$}
& {R-L$\downarrow$}
& {MMLU$\uparrow$}
& {Flu.$\uparrow$} \\
\midrule
Original & 74.76 & 85.14 & 46.39 & 4.03 \\
Retain   & 1.91  & 9.83 & 47.82 & 2.87  \\
\midrule
WHP & 23.55 & 17.93  & 40.40 & 2.52 \\
AS & \textbf{13.72} & \textbf{11.68} & 37.56 &  3.08 \\
ICUL & 38.47 & 32.49 & 39.86 & \textbf{3.60} \\
ULD & 27.20 & 22.97 & 39.24 & 3.36 \\
S{\small CANS}    & 23.36 & 20.24 & 40.10 & 3.19 \\
I{\small NNSTEER} & 19.92 & 17.72  & \underline{40.94} & 3.27 \\
ST$^{2}$U (Ours)  & \underline{14.61} & \underline{12.90}  & \textbf{42.34} & \underline{3.39} \\
\bottomrule
\end{tabular*}
}
\label{tab:main_muse_books}
\end{table}

\noindent$\bullet$ \textbf{Restricted Knowledge Re-entry.}
Lower benchmark forget scores do not establish that target knowledge remains inaccessible as generation continues.
Across backbones, ST\(^2\)U achieves R-Ent. pairs (RWKU/WMDP) of 19.20/13.76, 19.84/16.20, and 18.10/15.40, respectively, the lowest among the compared unlearning methods. The MUSE-Books experiments connect this persistence advantage to long-form copyright continuation. Under otherwise matched settings, removing correction history increases confirmed re-entry during continued decoding (Fig.~\ref{fig:obs}(b)), demonstrating the contribution of historical state to sustained suppression. Table~\ref{tab:main_muse_books} complements this mechanism-level comparison with end-task results: ST\(^2\)U achieves the second-lowest BLEU and ROUGE-L, together with the highest MMLU and second-highest fluency among the compared unlearning methods. Together, these results show that history-dependent control reduces re-entry in book continuation while providing a favorable forget--retain trade-off.
A qualitative example is given in Appendix.

\subsection{Parameter Sensitivity and Ablation}
\begin{figure}[t]
\centering
\includegraphics[width=1.0\linewidth]{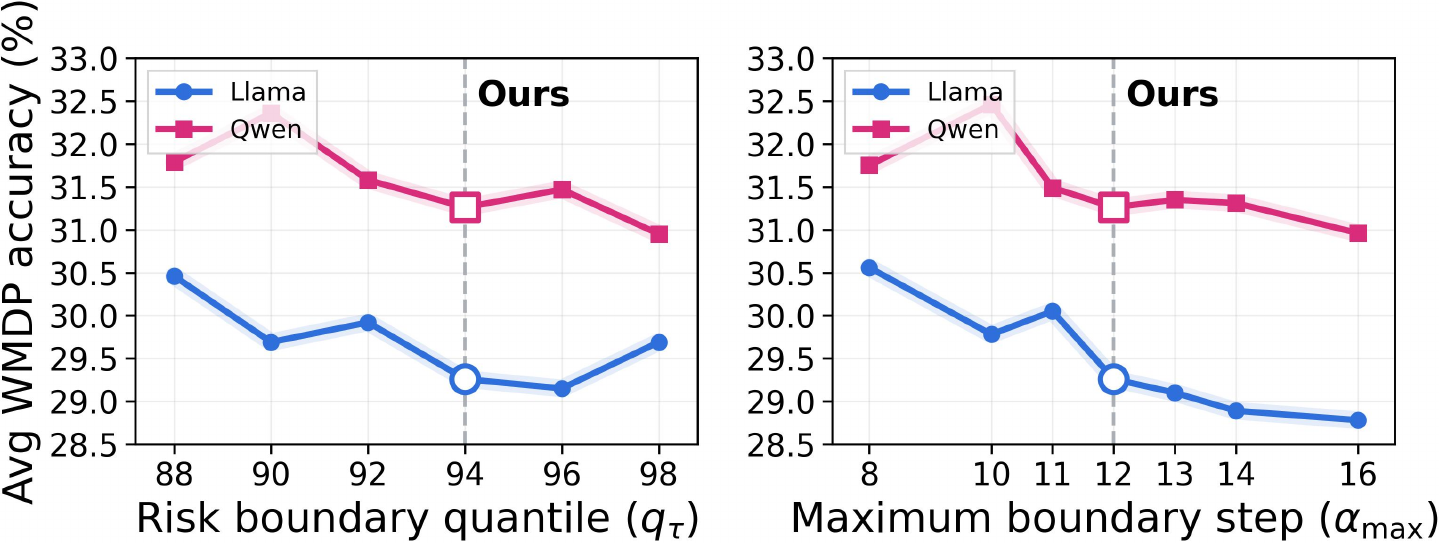}
\caption{Parameter sensitivity analysis on WMDP.} 
\label{fig:hyperparameter}

\end{figure}

\noindent$\bullet$ \textbf{Hyperparameter Sensitivity.} We vary one deployment hyperparameter at a time while keeping the learned target geometry fixed (see Fig.~\ref{fig:hyperparameter}). The risk-boundary quantile \(q_\tau\) determines when a hidden trajectory is treated as entering the restricted knowledge region. Sweeping it from 88\% to 98\% yields small, non-monotonic changes in WMDP accuracy, and the best-performing point differs across backbones. We therefore use one shared operating boundary instead of selecting a per-model oracle threshold. The maximum boundary step \(\alpha_{\max}\) caps the local closure strength. If the cap is too small, the controller may not move risky states past the boundary. Larger caps permit more aggressive corrections, trading retention for marginal WMDP reduction. We use \(\alpha_{\max}=12\) as a practical operating point: larger tested caps yield only modest additional WMDP reductions while lowering MMLU accuracy by 3.40--6.58 percentage points relative to this setting.

\noindent$\bullet$ \textbf{Ablation of Control Components.} 
Table~\ref{tab:component_ablation} evaluates the contributions of correction history, risk-triggered control, and residual preservation. The history ablation resets only controller memory between positions while retaining the backbone cache and token-wise intervention. Removing this memory increases R-Ent. despite improving MMLU by 1.52 percentage points, highlighting the importance of explicit correction history beyond backbone context. Correcting every token (w/o risk gate) increases average WMDP accuracy by 13.09 percentage points and R-Ent. by 32.99 percentage points while reducing MMLU. These results demonstrate that selective risk triggering improves suppression and retained utility. Finally, removing residual preservation causes the largest MMLU decline, indicating that retaining the orthogonal residual protects general utility during target-coordinate editing.

\begin{table}[h]
\centering
\caption{Ablation on Avg. WMDP (Llama-3.1-8B-Instruct).}
\resizebox{\columnwidth}{!}{
\begin{tabular}{lccc}
\toprule
\textbf{Variant}
& \textbf{WMDP$\downarrow$}
& \textbf{MMLU$\uparrow$}
& \textbf{R-Ent.$\downarrow$} \\
\midrule
Full ST$^{2}$U
& \textbf{29.26}
& {60.58}
& \textbf{13.76} \\

{w/o history}
& 37.23
& \textbf{62.10}
& 37.50 \\

w/o risk gate
& 42.35
& 55.55
& 46.75 \\

w/o residual preservation
& 35.45
& 52.79
& 18.75 \\
\bottomrule
\end{tabular}
}
\label{tab:component_ablation}
\end{table}

\noindent$\bullet$ \textbf{Effect of Correction History.}
Table~\ref{tab:history-diagnostics} evaluates session-level target leakage and target-response intervention magnitude in multi-turn WMDP dialogues using Llama-3.1-8B-Instruct. Each session contains five distinct MMLU questions and one WMDP target query, with the target query inserted at a randomly selected position. Unlike R-Ent., which counts reappearance after an initial successful intervention within a single generation, session-level output leakage measures whether confirmed target content appears anywhere in the dialogue, regardless of prior closure. Output length and intervention statistics are computed only for the WMDP target response. Edit magnitude is the norm of the final hidden-state change at each decoding position. Mean nonzero edit averages these norms over corrected positions, while total edit sums them per target response before averaging across sessions. Removing correction history raises output leakage from 15.72\% to 40.38\%. Mean nonzero edit remains similar (2.18 with history versus 2.24 without), while mean total edit decreases from 27.52 to 12.85 with history. Target responses average 37.4 versus 44.7 tokens, respectively. The ratio of mean total edit to mean response length also decreases from 0.616 to 0.344, indicating that shorter responses alone do not explain the lower cumulative perturbation. These results support the effectiveness of correction history for target suppression in multi-turn contexts with lower intervention cost.

\begin{table}[t]
\caption{Correction history output and intervention diagnostics on multi-turn WMDP dialogues (Llama-3.1-8B-Instruct).}
\centering
\resizebox{\columnwidth}{!}{
\begin{tabular}{lcccc}
\toprule
\textbf{\shortstack{Variant \\ \-}} & \textbf{\shortstack{Output\\leakage (\%) $\downarrow$}} & \textbf{\shortstack{Output\\length (tokens)}} & \textbf{\shortstack{Mean\\nonzero edit}} & \textbf{\shortstack{Total\\edit}} \\
\midrule
Full ST$^2$U & 15.72 & 37.4 & 2.18 & 12.85 \\
{w/o history} & 40.38 & 44.7 & 2.24 & 27.52 \\
\bottomrule
\end{tabular}
}
\label{tab:history-diagnostics}
\end{table}

\section{Further Analysis}

\subsection{Robustness to Extraction Attacks}

The results in Table~\ref{tab:main_rwku_wmdp} assess forgetting through benchmark probes. We extend this evaluation to recovery attempts using hidden-state perturbations and black-box extraction.
ICUL is evaluated through black-box extraction because the prompt-based intervention does not expose a comparable hidden-state control surface.

\noindent$\bullet$ \textbf{Limited White-box Attacks.}
The attacker knows the backbone and defense family and can inject low-rank hidden-state perturbations at the defended layer, but cannot disable runtime hooks or overwrite the controller state. We apply $\mathbf h_t^{\mathrm{atk}}=\mathbf h_t+\lambda_a\mathbf U_a\mathbf U_a^\top\mathbf h_t$, where $\mathbf U_a$ spans a method-specific attack subspace estimated from the defense's exposed intervention geometry and $\lambda_a$ controls perturbation strength. Recovery and $\Delta$MMLU denote the attacked-minus-clean changes in Avg. WMDP and MMLU accuracy.

\noindent$\bullet$ \textbf{Black-box Extraction.}
Prompt variation (PV) modifies the target request through paraphrasing, masking, or formatting changes. Multi-turn delayed reveal (MT) introduces benign conversational context before requesting target knowledge. Sampling (S@32) draws 32 continuations per prompt. We report R-Ent. under each setting.

Table~\ref{tab:attack_robustness} highlights different failure patterns across methods. RMU's limited white-box recovery accompanies substantial utility degradation, making its low recovery less indicative of selective protection. SEGUE exhibits moderate white-box robustness with 8.10 points of WMDP recovery and a 3.70-point MMLU drop, falling between parameter-updating RMU and activation-steering SCANS in both defense strength and utility retention. SCANS shows the opposite pattern to RMU: target knowledge becomes more recoverable while MMLU improves, suggesting that its suppression can be weakened without damaging general capabilities. ST\(^2\)U achieves the smallest recovery (+1.16 points) with a modest MMLU loss ($-$1.92 points). Under black-box extraction, ICUL remains highly recoverable, while the stronger exposure of RMU and SCANS under delayed requests and repeated sampling suggests that suppression depends on the interaction context and decoding path. ST\(^2\)U retains the lowest R-Ent. across all settings, consistent with the role of historical correction state in sustaining suppression beyond an individual intervention.

\begin{table}[t]
\caption{Attack robustness (Llama-3.1-8B-Instruct). White-box: percentage points. Black-box: \%.}
\centering
\begin{tabular}{lccccc}
\toprule
\multirow{2}{*}{\textbf{Method}} & \multicolumn{2}{c}{\textbf{White-box (Avg. WMDP)}} & \multicolumn{3}{c}{\textbf{Black-box (R-Ent.)}} \\
\cmidrule(lr){2-3}\cmidrule(lr){4-6}
 & Recovery $\downarrow$ & $\Delta$MMLU $\uparrow$ & PV $\downarrow$ & MT $\downarrow$ & S@32 $\downarrow$ \\
\midrule
RMU & 4.87 & $-17.32$ & 27.6 & 32.4 & 38.0 \\
ICUL & -- & -- & 86.2 & 89.1 & 87.1 \\
SCANS & 15.84 &\textbf{$+$1.28} & 59.2 & 66.5 & 64.9 \\
SEGUE & 8.10 & $-$3.70 & 39.5 & 45.7 & 48.5 \\
ST$^2$U & \textbf{1.16} & $-1.92$ & \textbf{14.6} & \textbf{17.6} & \textbf{19.3} \\
\bottomrule
\end{tabular}
\label{tab:attack_robustness}
\end{table}

\begin{figure}[t]
\centering
\includegraphics[width=1.0\linewidth]{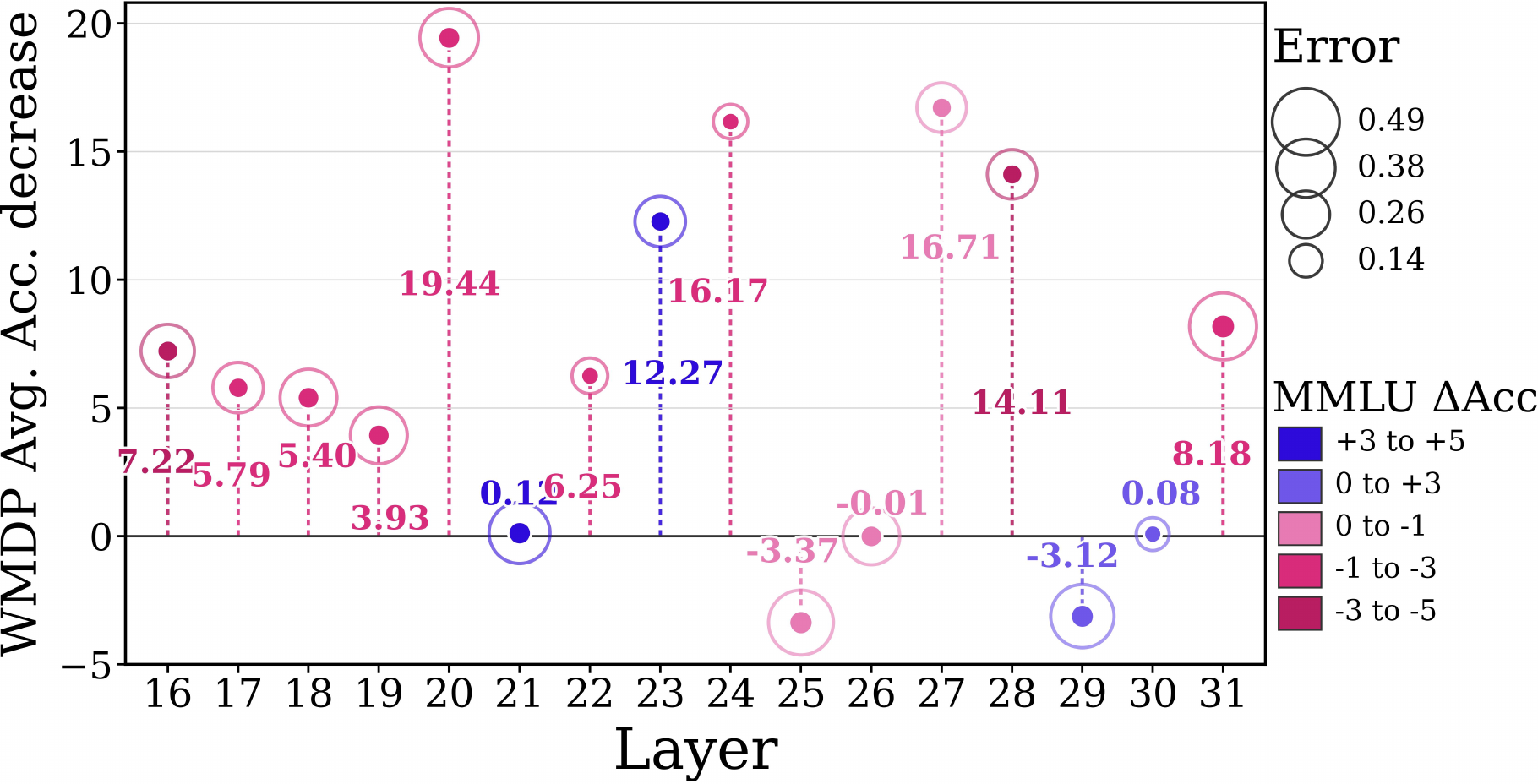}
\caption{Layer-wise gains from low-dimensional invertible coordinates over same-layer direct hidden steering on Llama-3.1-8B. Positive values denote additional Avg. WMDP accuracy reduction. Colors indicate MMLU accuracy changes, and bubble sizes encode the standard deviation across runs.}
\label{fig:layer_robustness}
\end{figure}

\subsection{Effectiveness of Invertible Coordinates}
\noindent$\bullet$ \textbf{Layer-wise Comparison.}
Figure~\ref{fig:layer_robustness} compares same-layer direct hidden steering with steering augmented by low-dimensional invertible coordinates across layers 16--31 of Llama-3.1-8B. The vertical axis measures the additional reduction in Avg. WMDP accuracy, with larger positive values indicating stronger forgetting over direct steering. Of the 16 layers, 13 improve, 10 gain at least 5 WMDP points, and 6 exceed 8 points, showing that the benefit is not confined to one selected layer. Layers 20, 23, 24, 27, and 31 combine large forgetting gains with at most 2.5 points of MMLU loss. Layer 23 is particularly illustrative, reducing WMDP accuracy by an additional 12.27 points while improving MMLU by 4.72 points. 
These results show that the invertible coordinates' benefit extends across multiple intervention layers, although the forget--retain trade-off remains layer-dependent. At layer 29, which is used in our main experiments, invertible coordinates improve MMLU by 1.80 points while increasing WMDP accuracy by 3.12 points relative to direct steering. Layer selection therefore considers the absolute forget--retain operating point, not only the coordinate gain.

\begin{figure}[t]
\centering
\includegraphics[width=\linewidth]{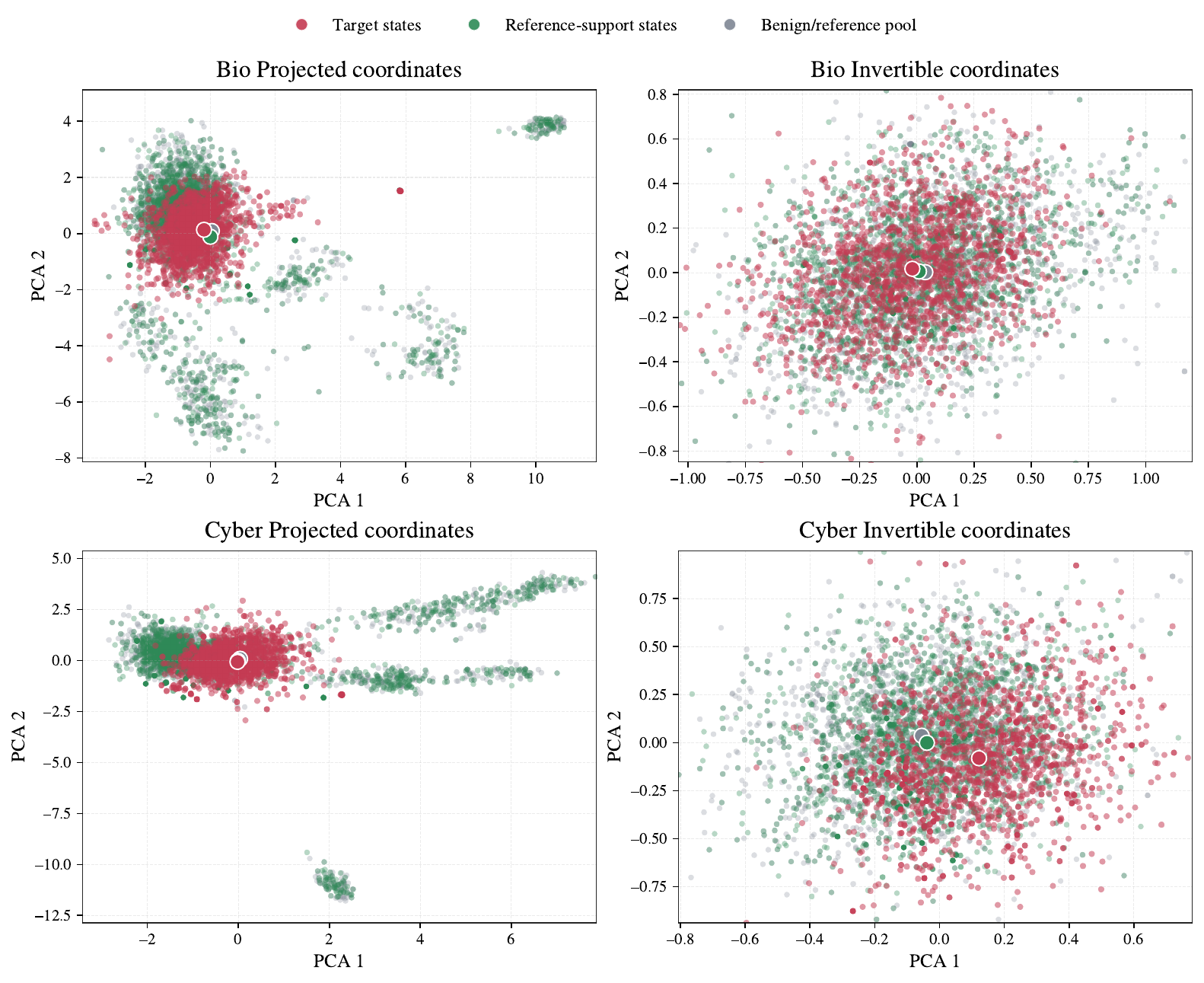}
\caption{PCA views of layer-23 states on Llama-3.1-8B-Instruct. Rows correspond to Bio and Cyber. Columns show projected and invertibly transformed coordinates. Colors distinguish target states, reference-support states, and the benign reference pool. Axis ranges differ across panels.}
\label{fig:coordinate_pca}
\end{figure}

\noindent$\bullet$ \textbf{Representation Geometry.}
Figure~\ref{fig:coordinate_pca} illustrates the transformation at layer 23. The projected Bio and Cyber representations contain elongated structures and dispersed groups, which become more compact after the invertible map. This reshaping preserves an analytic inverse, allowing edited coordinates to be mapped back to the original projected representation, and provides a more regular coordinate geometry for estimating target-specific densities and following their risk boundaries. The gains over direct steering at most layers show that the choice of coordinates matters beyond where the intervention is applied. Additional risk-plane visualization and geometric analysis are presented in the Appendix.

\subsection{Deployment Cost}
\begin{figure}[t]
\centering
\includegraphics[width=1.0\linewidth]{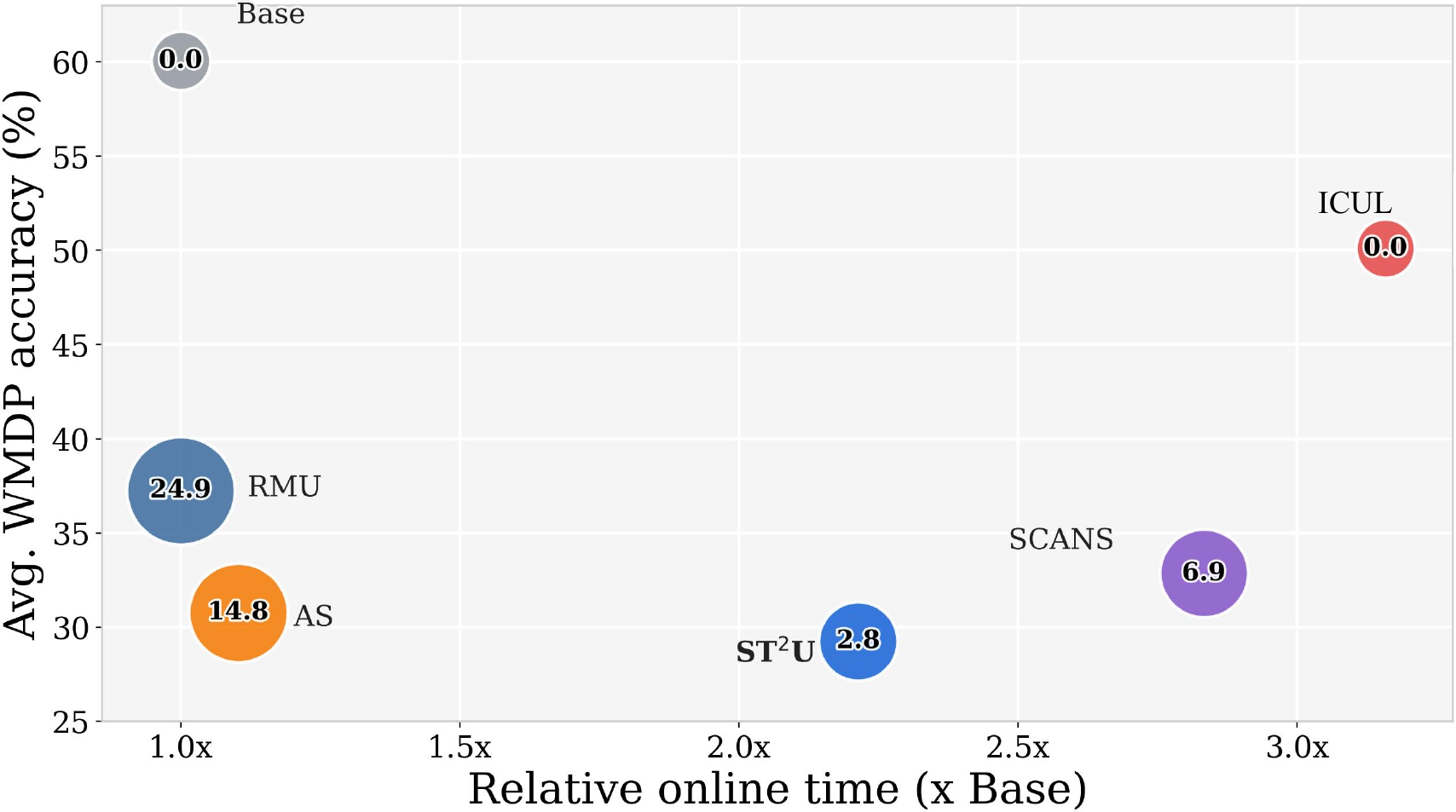}
\caption{Relative deployment cost (Llama-3.1-8B-Instruct). Bubble labels give offline preparation time in minutes.}
\label{fig:deployment_cost}
\end{figure}

Figure~\ref{fig:deployment_cost} compares deployment efficiency, where bubble area represents measured offline preparation time in minutes and the horizontal axis reports online latency relative to Base. ST\(^{2}\)U achieves the lowest Avg. WMDP accuracy (29.26) and the highest retained MMLU accuracy (60.58) among all unlearning methods, with only 2.8 minutes of offline preparation. Although its \(2.21\times\) online latency exceeds AS and RMU, these parameter-updating methods require 14.8 and 24.9 offline minutes while producing weaker forgetting and retention. ICUL eliminates offline preparation but is slower online (\(3.16\times\)) and retains substantially more restricted knowledge (50.10). SEGUE incurs minimal offline overhead (1.1 min) as a decoding-level test-time unlearning method, but its per-token entropy-guided distribution recalibration results in higher online latency than ST\(^{2}\)U and a less competitive forget--retain trade-off. Moreover, ST\(^{2}\)U strictly outperforms SCANS across offline time, online latency, WMDP, and MMLU. Overall, ST\(^{2}\)U provides the best forget--retain operating point in this comparison at a moderate deployment cost.

\section{Conclusion}
We reframe test-time unlearning as maintaining restricted knowledge suppression throughout autoregressive generation: a locally successful correction does not ensure that subsequent states remain below the risk boundary. We introduce ST$^{2}$U, which constructs target-specific risk geometries in low-dimensional invertible coordinates, preserves the orthogonal complement, and carries correction history through a fixed-dimensional state. By combining boundary correction, contextual anchoring, and tangential state guidance, ST$^{2}$U connects local risk reduction with persistent, context-aware control across decoding steps. The estimated boundary is calibrated on finite data, leaving its coverage under broader contextual shifts open to further study. More broadly, this perspective connects unlearning to stateful inference and constrained dynamical systems, highlighting runtime memory as a resource for regulating knowledge access during generation. An open question is what information about past corrections a fixed-dimensional state must retain to reduce future boundary violations as context evolves. Future work will investigate the relationship between boundary generalization, state capacity, and knowledge re-entry, and develop adaptive intervention schedules that reduce inference overhead while preserving forgetting effectiveness and non-target utility.

\bibliography{aaai2027}

@article{pan2025spikingbrain,
  author       = {Yuqi Pan and
                  Yupeng Feng and
                  Jinghao Zhuang and
                  Siyu Ding and
                  Han Xu and
                  Zehao Liu and
                  Bohan Sun and
                  Yuhong Chou and
                  Xuerui Qiu and
                  Anlin Deng and
                  Anjie Hu and
                  Shurong Wang and
                  Peng Zhou and
                  Man Yao and
                  Jibin Wu and
                  Jian Yang and
                  Bo Xu and
                  Guoqi Li},
  title        = {SpikingBrain: Spiking Brain-inspired Large Models},
  journal      = {Trans. Mach. Learn. Res.},
  volume       = {2026},
  year         = {2026}
}

@inproceedings{chen2026alter,
  author       = {Xunlei Chen and
                  Jinyu Guo and
                  Yuang Li and
                  Zhaokun Wang and
                  Yi Gong and
                  Jie Zou and
                  Jiwei Wei and
                  Wenhong Tian},
  editor       = {Sven Koenig and
                  Chad Jenkins and
                  Matthew E. Taylor},
  title        = {{ALTER:} Asymmetric LoRA for Token-Entropy-Guided Unlearning of LLMs},
  booktitle    = {Fortieth {AAAI} Conference on Artificial Intelligence, Thirty-Eighth
                  Conference on Innovative Applications of Artificial Intelligence,
                  Sixteenth Symposium on Educational Advances in Artificial Intelligence,
                  {AAAI} 2026, Singapore, January 20-27, 2026},
  pages        = {35366--35374},
  publisher    = {{AAAI} Press},
  year         = {2026}
}

@inproceedings{NEURIPS2025_c9826b9e,
 author = {Zharmagambetov, Arman and Guo, Chuan and Evtimov, Ivan and Pavlova, Maya and Salakhutdinov, Ruslan and Chaudhuri, Kamalika},
 booktitle = {Advances in Neural Information Processing Systems},
 doi = {10.52202/085713-4609},
 editor = {D. Belgrave and C. Zhang and H. Lin and R. Pascanu and P. Koniusz and M. Ghassemi and N. Chen},
 pages = {},
 publisher = {Curran Associates, Inc.},
 title = {AgentDAM: Privacy Leakage Evaluation for Autonomous Web Agents},
 url = {https://proceedings.neurips.cc/paper_files/paper/2025/file/c9826b9ea5e1b49b256329934a578d83-Paper-Datasets_and_Benchmarks_Track.pdf},
 volume = {38, Main Conference},
 year = {2025}
}

@inproceedings{yuan2025query,
  title={Query understanding in llm-based conversational information seeking},
  author={Yuan, Yifei and Abbasiantaeb, Zahra and Aliannejadi, Mohammad and Deng, Yang},
  booktitle={Proceedings of the 48th International ACM SIGIR Conference on Research and Development in Information Retrieval},
  pages={4098--4101},
  year={2025}
}

@book{scott2015multivariate,
  title={Multivariate density estimation: theory, practice, and visualization},
  author={Scott, David W},
  year={2015},
  publisher={John Wiley \& Sons}
}

@InProceedings{pmlr-v306-sun26d,
  title = 	 {Anatomy of Massive Activations and Attention Sinks},
  author =       {Sun, Shangwen and Canziani, Alfredo and Lecun, Yann and Zhu, Jiachen},
  booktitle = 	 {Proceedings of the 43rd International Conference on Machine Learning},
  pages = 	 {116443--116463},
  year = 	 {2026},
  volume = 	 {306},
  publisher =    {PMLR},
  url = 	 {https://proceedings.mlr.press/v306/sun26d.html},
}

@inproceedings{guo2025forgetting,
  title={Forgetting through transforming: Enabling federated unlearning via class-aware representation transformation},
  author={Guo, Qi and Tian, Zhen and Yao, Minghao and Qi, Saiyu and Qi, Yong and Liu, Bingyi},
  booktitle={2025 IEEE/CVF International Conference on Computer Vision (ICCV)},
  pages={1474--1483},
  year={2025},
  organization={IEEE}
}

@inproceedings{shao2026baldro,
  title={Baldro: A distributionally robust optimization based framework for large language model unlearning},
  author={Shao, Pengyang and Zhai, Naixin and Chen, Lei and Yang, Yonghui and Zhu, Fengbin and Yang, Xun and Wang, Meng},
  booktitle={Proceedings of the ACM Web Conference 2026},
  pages={8874--8884},
  year={2026}
}

@article{cao2024rwku,
  title={Rwku: Benchmarking real-world knowledge unlearning for large language models},
  author={Cao, Pengfei and Wang, Chenhao and He, Zhitao and Yuan, Hongbang and Li, Jiachun and Chen, Yubo and Liu, Kang and Zhao, Jun and others},
  journal={Advances in Neural Information Processing Systems},
  volume={37},
  pages={98213--98263},
  year={2024}
}

@inproceedings{li2024wmdp,
  title={The WMDP benchmark: measuring and reducing malicious use with unlearning},
  author={Li, Nathaniel and Pan, Alexander and Gopal, Anjali and Yue, Summer and Berrios, Daniel and Gatti, Alice and Li, Justin D and Dombrowski, Ann-Kathrin and Goel, Shashwat and Mukobi, Gabriel and others},
  booktitle={Proceedings of the 41st International Conference on Machine Learning},
  pages={28525--28550},
  year={2024}
}

@InProceedings{Gong_2026_CVPR,
  author    = {Gong, Qinghui and Yang, Xue and Chen, Xunlei and Lai, Jinshan and Meng, Hua and Tang, Xiaohu},
  title     = {FedOrtho: Efficient Federated Unlearning Via Orthogonal Convolution and Adaptive Soft Pruning},
  booktitle = {Proceedings of the IEEE/CVF Conference on Computer Vision and Pattern Recognition (CVPR) Findings},
  month     = {June},
  year      = {2026},
  pages     = {8009--8018}
}

@article{yu2025unierase,
  title={UniErase: Unlearning Token as a Universal Erasure Primitive for Language Models},
  author={Yu, Miao and Lin, Liang and Zhang, Guibin and Li, Xinfeng and Fang, Junfeng and Zhang, Ningyu and Wang, Kun and Wang, Yang},
  journal={arXiv preprint arXiv:2505.15674},
  year={2025},
  url={https://arxiv.org/abs/2505.15674}
}

@inproceedings{wang2026cap,
  title={CAP: Controllable Alignment Prompting for Unlearning in {LLM}s},
  author={Wang, Zhaokun and Guo, Jinyu and Pu, Jingwen and Pu, Hongli and Yang, Meng and Chen, Xunlei and Ou, Jie and Li, Wenyi and Luo, Guangchun and Tian, Wenhong},
  booktitle={Proceedings of the 64th Annual Meeting of the Association for Computational Linguistics (Volume 1: Long Papers)},
  year={2026}
}

@article{eldan2023s,
  title={Who's Harry Potter? Approximate Unlearning in LLMs},
  author={Eldan, Ronen and Russinovich, Mark},
  journal={arXiv preprint arXiv:2310.02238},
  year={2023}
}

@article{ji2024reversing,
  title={Reversing the forget-retain objectives: An efficient llm unlearning framework from logit difference},
  author={Ji, Jiabao and Liu, Yujian and Zhang, Yang and Liu, Gaowen and Kompella, Ramana R and Liu, Sijia and Chang, Shiyu},
  journal={Advances in Neural Information Processing Systems},
  volume={37},
  pages={12581--12611},
  year={2024}
}

@inproceedings{shi2025muse,
  title={Muse: Machine unlearning six-way evaluation for language models},
  author={Shi, Weijia and Lee, Jaechan and Huang, Yangsibo and Malladi, Sadhika and Zhao, Jieyu and Holtzman, Ari and Liu, Daogao and Zettlemoyer, Luke and Smith, Noah and Zhang, Chiyuan},
  booktitle={International Conference on Learning Representations},
  volume={2025},
  pages={27797--27818},
  year={2025}
}

@inproceedings{xu2025obliviate,
  title={OBLIVIATE: Robust and Practical Machine Unlearning for Large Language Models},
  author={Xu, Xiaoyu and Du, Minxin and Ye, Qingqing and Hu, Haibo},
  booktitle={Proceedings of the 2025 Conference on Empirical Methods in Natural Language Processing},
  pages={3696--3715},
  year={2025}
}

@article{hendrycks2020measuring,
  title={Measuring massive multitask language understanding},
  author={Hendrycks, Dan and Burns, Collin and Basart, Steven and Zou, Andy and Mazeika, Mantas and Song, Dawn and Steinhardt, Jacob},
  journal={arXiv preprint arXiv:2009.03300},
  year={2020}
}

@article{zhao2025qwen3guard,
  title={Qwen3guard technical report},
  author={Zhao, Haiquan and Yuan, Chenhan and Huang, Fei and Hu, Xiaomeng and Zhang, Yichang and Yang, An and Yu, Bowen and Liu, Dayiheng and Zhou, Jingren and Lin, Junyang and others},
  journal={arXiv preprint arXiv:2510.14276},
  year={2025}
}

@inproceedings{pu2026decoding,
  title={Decoding-Unlearning: Fact Forgetting via Entropy-Guided Inference},
  author={Pu, Jingwen and Shi, Mingjun and Ren, Xinrui and Wang, Yizhe and Zhang, Xinyu and Wang, Zhaokun and She, Kun},
  booktitle={Proceedings of the 64th Annual Meeting of the Association for Computational Linguistics (Volume 1: Long Papers)},
  pages={39834--39860},
  year={2026}
}

@inproceedings{cao2025scans,
  title={Scans: Mitigating the exaggerated safety for llms via safety-conscious activation steering},
  author={Cao, Zouying and Yang, Yifei and Zhao, Hai},
  booktitle={Proceedings of the AAAI Conference on Artificial Intelligence},
  volume={39},
  number={22},
  pages={23523--23531},
  year={2025}
}

@article{nguyen2026beyond,
  title={Beyond Linear Activation Steering: Invertible Latent Transformations for Controlling LLM Behavior},
  author={Nguyen, Tuc and Le, Thai},
  journal={arXiv preprint arXiv:2606.08454},
  year={2026}
}

@inproceedings{lee2026direct,
  title={Direct Token Optimization: A Self-Contained Approach to Large Language Model Unlearning},
  author={Lee, Hong Kyu and Liu, Ruixuan and Xiong, Li},
  booktitle={Findings of the Association for Computational Linguistics: ACL 2026},
  pages={42083--42100},
  year={2026}
}

@article{sanyal2025agents,
  title={Agents are all you need for LLM unlearning},
  author={Sanyal, Debdeep and Mandal, Murari},
  journal={arXiv preprint arXiv:2502.00406},
  year={2025}
}

@inproceedings{bhaila2025soft,
  title={Soft prompting for unlearning in large language models},
  author={Bhaila, Karuna and Van, Minh-Hao and Wu, Xintao},
  booktitle={Proceedings of the 2025 Conference of the Nations of the Americas Chapter of the Association for Computational Linguistics: Human Language Technologies (Volume 1: Long Papers)},
  pages={4046--4056},
  year={2025}
}

@article{wang2026fupareto,
  title={FUPareto: Bridging the Forgetting-Utility Gap in Federated Unlearning via Pareto Augmented Optimization},
  author={Wang, Zeyan and Liu, Zhengmao and Cai, Yongxin and Li, Chi and Tang, Xiaoying and Chen, Jingchao and Pan, Zibin and Qiu, Jing},
  journal={arXiv preprint arXiv:2602.01852},
  year={2026},
  url={https://arxiv.org/abs/2602.01852}
}

@article{liu2025rethinking,
  title={Rethinking machine unlearning for large language models},
  author={Liu, Sijia and Yao, Yuanshun and Jia, Jinghan and Casper, Stephen and Baracaldo, Nathalie and Hase, Peter and Yao, Yuguang and Liu, Chris Yuhao and Xu, Xiaojun and Li, Hang and others},
  journal={Nature Machine Intelligence},
  volume={7},
  number={2},
  pages={181--194},
  year={2025},
  publisher={Nature Publishing Group UK London}
}

@article{liu2024large,
  title={Large language model unlearning via embedding-corrupted prompts},
  author={Liu, Chris Y and Wang, Yaxuan and Flanigan, Jeffrey and Liu, Yang},
  journal={Advances in Neural Information Processing Systems},
  volume={37},
  pages={118198--118266},
  year={2024}
}

@inproceedings{li2026cipo,
  title={CiPO: Counterfactual Unlearning for Large Reasoning Models through Iterative Preference Optimization},
  author={Li, Junyi and Chen, Yongqiang and Ding, Ningning},
  booktitle={Proceedings of the 64th Annual Meeting of the Association for Computational Linguistics (Volume 1: Long Papers)},
  pages={3152--3170},
  year={2026}
}

@inproceedings{tutek2025measuring,
  title={Measuring chain of thought faithfulness by unlearning reasoning steps},
  author={Tutek, Martin and Chaleshtori, Fateme Hashemi and Marasovi{\'c}, Ana and Belinkov, Yonatan},
  booktitle={Proceedings of the 2025 Conference on Empirical Methods in Natural Language Processing},
  pages={9946--9971},
  year={2025}
}

@inproceedings{wang2025towards,
  title={Towards lifecycle unlearning commitment management: Measuring sample-level unlearning completeness},
  author={Wang, Cheng-Long and Li, Qi and Xiang, Zihang and Cao, Yinzhi and Wang, Di},
  booktitle={34th USENIX Security Symposium (USENIX Security 25)},
  pages={6481--6500},
  year={2025}
}

@inproceedings{xu2025relearn,
  title={Relearn: Unlearning via learning for large language models},
  author={Xu, Haoming and Zhao, Ningyuan and Yang, Liming and Zhao, Sendong and Deng, Shumin and Wang, Mengru and Hooi, Bryan and Oo, Nay and Chen, Huajun and Zhang, Ningyu},
  booktitle={Proceedings of the 63rd Annual Meeting of the Association for Computational Linguistics (Volume 1: Long Papers)},
  pages={5967--5987},
  year={2025}
}

@inproceedings{li2026cross,
  title={Cross-modal unlearning via influential neuron path editing in multimodal large language models},
  author={Li, Kunhao and Li, Wenhao and Wu, Di and Yang, Lei and Bai, Jun and Jia, Ju and Xue, Jason},
  booktitle={Proceedings of the AAAI Conference on Artificial Intelligence},
  volume={40},
  number={42},
  pages={35589--35597},
  year={2026}
}

@article{shen2026llm,
  title={LLM unlearning via neural activation redirection},
  author={Shen, William and Qiu, Xinchi and Kurmanji, Meghdad and Iacob, Alexandru-Andrei and Sani, Lorenzo and Chen, Yihong and Cancedda, Nicola and Lane, Nicholas},
  journal={Advances in Neural Information Processing Systems},
  volume={38},
  pages={44253--44290},
  year={2026}
}

@inproceedings{thudi2022unrolling,
  title={Unrolling sgd: Understanding factors influencing machine unlearning},
  author={Thudi, Anvith and Deza, Gabriel and Chandrasekaran, Varun and Papernot, Nicolas},
  booktitle={EuroS\&P 2022},
  pages={303--319},
  year={2022},
  organization={IEEE}
}

@article{zhang2025price,
  title={The price of unlearning: identifying unlearning risk in edge computing},
  author={Zhang, Lefeng and Zhu, Tianqing and Xiong, Ping and Zhou, Wanlei},
  journal={ACM Transactions on Multimedia Computing, Communications and Applications},
  volume={21},
  number={8},
  pages={1--23},
  year={2025},
  publisher={ACM New York, NY}
}

@inproceedings{ranjan2026razor,
  title={Razor: Ratio-aware layer editing for targeted unlearning in vision transformers and diffusion models},
  author={Ranjan, Ravi and Grover, Utkarsh and Lin, Xiaomin and Polyzou, Agoritsa},
  booktitle={Proceedings of the IEEE/CVF Conference on Computer Vision and Pattern Recognition},
  pages={7998--8008},
  year={2026}
}

@inproceedings{pawelczyk2024context,
  title={In-Context Unlearning: Language Models as Few-Shot Unlearners},
  author={Pawelczyk, Martin and Neel, Seth and Lakkaraju, Himabindu},
  booktitle={International Conference on Machine Learning},
  pages={40034--40050},
  year={2024},
  organization={PMLR}
}

@article{yang2025qwen3,
  title={Qwen3 technical report},
  author={Yang, An and Li, Anfeng and Yang, Baosong and Zhang, Beichen and Hui, Binyuan and Zheng, Bo and Yu, Bowen and Gao, Chang and Huang, Chengen and Lv, Chenxu and others},
  journal={arXiv preprint arXiv:2505.09388},
  year={2025}
}

@article{tan2026wisdom,
  title={Wisdom is knowing what not to say: Hallucination-free LLMs unlearning via attention shifting},
  author={Tan, Chenchen and Qu, Youyang and Li, Xinghao and Zhang, Hui and Cui, Shujie and Chen, Cunjian and Gao, Longxiang},
  journal={Advances in Neural Information Processing Systems},
  volume={38},
  pages={99138--99166},
  year={2026}
}

@inproceedings{grattafiori2024llama,
  title={The Llama 3 herd of models},
  author={Grattafiori, Aaron and Dubey, Abhimanyu and Jauhri, Abhinav and Pandey, Abhinav and Kadian, Abhishek and Al-Dahle, Ahmad and Letman, Aiesha and Mathur, Akhil and Schelten, Alan and Vaughan, Alex and others},
  booktitle={Neural Information Processing Systems},
  year={2024},
  organization={Curran Associates}
}

\appendix

\end{document}